\documentclass{article}

\usepackage[utf8]{inputenc} 
\usepackage[T1]{fontenc}    
\usepackage{fullpage}
\usepackage[round]{natbib}

\usepackage{microtype}
\usepackage{graphicx}
\usepackage{subfigure}
\usepackage{booktabs} 
\usepackage{color}
\usepackage[table]{xcolor}  %
\usepackage{caption}

\usepackage{amsmath}
\usepackage{amssymb}
\usepackage{mathtools}
\usepackage{amsthm}

\usepackage{amsfonts}       
\usepackage{bbold}
\usepackage{mathtools}
\usepackage{tikz}
\usetikzlibrary{arrows}
\usetikzlibrary{calc}
\usetikzlibrary{fit}
\usepackage{multirow}

\definecolor{puorange}{rgb}{0.80,0.20,0}
\definecolor{bluegray}{rgb}{0.04,0,0.7}
\definecolor{greengray}{rgb}{0.05,0.50,0.15}
\definecolor{darkbrown}{rgb}{0.40,0.2,0.05}
\definecolor{darkcyan}{rgb}{0,0.4,1}
\definecolor{black}{rgb}{0,0,0}
\definecolor{grey}{rgb}{0.93,0.93,0.93}
\definecolor{royalazure}{rgb}{0.0, 0.22, 0.66}

\usepackage[colorlinks,citecolor=bluegray,linkcolor=darkbrown,urlcolor=blue,breaklinks]{hyperref}
\usepackage[capitalize,noabbrev]{cleveref}

\include{math_commands}

\theoremstyle{plain}

\theoremstyle{definition}

\theoremstyle{remark}

\title{Predicting on the Edge: \\
            Identifying Where a Larger Model Does Better}
\author{Taman Narayan \qquad Heinrich Jiang \qquad Sen Zhao \qquad Sanjiv Kumar \vspace{0.3cm} \\
Google Research}
\date{}

\begin{document}

\maketitle

\begin{abstract}
Much effort has been devoted to making large and more accurate models, but relatively little has been put into understanding which examples are benefiting from the added complexity. In this paper, we demonstrate and analyze the surprisingly tight link between a model’s predictive uncertainty on individual examples and the likelihood that larger models will improve prediction on them. Through extensive numerical studies on the T5 encoder-decoder architecture, we show that large models have the largest improvement on examples where the small model is most uncertain. On more certain examples, even those where the small model is not particularly accurate, large models are often unable to improve at all, and can even perform worse than the smaller model. Based on these findings, we show that a \emph{switcher model} which defers examples to a larger model when a small model is uncertain can achieve striking improvements in performance and resource usage. We also explore committee-based uncertainty metrics that can be more effective but less practical.
\end{abstract}

\section{Introduction}
\label{introduction}

In recent years, it has been shown that dramatically increasing model size can lead to better performance on a variety of challenging problems including sequence-to-sequence language modeling \cite{brown2020language,raffel2019exploring}, image understanding \cite{iandola2014densenet,szegedy2017inception}, medical imaging \cite{mustafa2021supervised}, agriculture \cite{mishra2021analysis}, computational drug discovery \cite{hudson2021data}, self-driving \cite{fernandes2021point} and human activity recognition \cite{ronald2021isplinception}. Alongside the recent increased interest towards using, training, and deploying more powerful models-- to the limits of what is feasible with modern hardware-- has come ever-increasing requirements for increased computational resources. Thus, it is becoming ever more important to understand the trade-offs between the benefits of using larger models and the added computational complexity and cost, which has received surprisingly little treatment in the literature.

In this paper, we focus on one aspect of this issue. We present a simple yet effective method to identify examples on which larger models are likely to be able to improve prediction - and examples on which the larger model may get worse.  Knowing this can be useful in determining which examples should receive more compute resources during inference in order to improve overall efficiency of the system so that we can make accurate predictions using only the necessary amount of resources per example. Our findings thus may be of relevance to the area of {\it adaptive} inference \cite{nan2017adaptive,zhou2020bert}.

They may also be relevant to practitioners making decisions about potentially replacing an existing model with a more sophisticated model. Knowing for which examples there is likely to be improvements from introducing such added complexity can be helpful in making these decisions before investing resources. This is especially true in cases where performance must be stable, monitored, and {\it fair} across various population subgroups of the data, such as geographic regions \cite{cotter2019optimization} and demographic groups \cite{mehrabi2021survey}.
Moreover, it is often the case that labeled data is sparse and imbalanced \cite{schelter2018challenges} causing evaluation of the new model to be  expensive and challenging in both online and offline settings \cite{kossen2021active}; but access to demographic information is often available or can be easily inferred \cite{zhong2015you}. Our method thus has a welcome property that it doesn't require labels to determine whether prediction is likely to be improved with a larger model.

We summarize our contributions as follows.
\begin{itemize}
    \item We show that a smaller model's uncertainty can identify examples on which a larger model is likely to be able to improve and those on which is it not. It can also, in some cases, identify examples on which a larger model is likely to do worse. Our findings hold across various pairings of small and large models.
    \item We show that a switcher model (in which a smaller model's uncertainty decides whether or not to defer prediction to a larger model) can lead to substantial computational savings and even performance improvements relative to using the larger model alone. This last case, when a switcher model outperforms the larger model, is more likely when the small and large models are relatively close in terms of performance.
    \item We show that averaging the uncertainties of various re-trainings of the smaller model is less effective at identifying where larger models improve and decline over a particular model than using that model's uncertainty alone.
    \item We show that per-example re-training churn (i.e. model prediction disagreement) is strongly related both to a small model's uncertainty and a larger model's potential improvement. In fact, it can predict a larger model's potential improvement or regression even better than the uncertainty.
\end{itemize}

We also contextualize these findings, discuss their implications, and offer hypotheses and explanations. 

\section{Related Work}
\label{literature}

The most related study is that of \citet{zhong-etal-2021-larger}, who recently investigated where improvements from larger models came from. Like us, they found that larger language models are not uniformly better than smaller models on all test examples i.e. in one setting BERT-LARGE was worse than BERT-BASE on 4.5\% of the instances and better on 7\%. In our paper, we provide a simple method that can identify where a larger model does better leading to e.g. a way that can intelligently combine a larger model and a smaller model to achieve better performance and at the same time lower resource usage.

\citet{hendrycks-etal-2020-pretrained} and \citet{desai-durrett-2020-calibration} also explore when larger models perform better, but in the context of out-of-distribution performance. \cite{hendrycks-etal-2020-pretrained} found that larger models pre-trained on more diverse datasets are more robust to various distribution-shifts. However they do not explore on which examples improvements come from and in our setting we do not consider explicit distribution-shifts. \cite{desai-durrett-2020-calibration} showed that out-of-distribution calibration performance improves with pre-training.

\citet{pmlr-v119-sagawa20a} found that larger models (without pretraining) tends to perform worse on rare subgroups, which provides insight into when larger models can hurt performance. However their investigation did not consider pre-trained variants; moreover, their approach requires knowledge of such minority subgroups. 

\citet{brutzkus2019larger} provides an analysis for why larger models generalize better, but from a theoretical perspective. \citet{baldock2021deep} explore example difficult by using a $k$-NN based approach on the intermediate representation. \citet{toneva2018empirical} propose to quantiy example difficulty by the number of training iterations required to learn it. \citet{jiang2020characterizing} provide a statistical notion of example difficulty. In our work, the goal is not to quantify which examples are inherently difficult, but rather which examples' performance can be improved with a larger model.

Our paper also builds a connection between example churn and example difficulty, where churn is defined as the amount of disagreement in the model prediction from retrainings. Work in churn primarily focus on pointing out the surprising churn that arises from retraining \cite{sellam2021multiberts,d2020underspecification} and how to reduce such churn \cite{milani2016launch,cotter2019optimization,jiang2021churn,bahri2021locally,bhojanapalli2021reproducibility}. To our knowledge, this paper is the first to connect churn to example difficulty. 

Our paper also provides an interesting finding about model uncertainty in that the average uncertainty over retraining, which is known to be a stronger uncertainty estimate than the uncertainty of a single model \cite{parker2013ensemble,lakshminarayanan2016simple} turns out to perform worse in identifying examples in which the smaller model is likely to be right on. \citet{jiang2021bootstrapping} showed a similar conclusion when using uncertainties in the active learning context.  \citet{hendrycks2016baseline} found that using the model uncertainties is an effective baseline for identifying misclassified and out-of-distribution samples.

Finally, our work also relates to prior work that ensembles models of different sizes. \citet{rawat2021summon} consider modifying a standard distillation framework to account for examples that are difficult and allow for the teacher model to be used at evaluation time. \citet{wang2018idk} consider a multi-model cascade in which models can decide whether to predict a class or a dummy class that indicates that they don't know. Like them, we share a goal of using simpler models when possible and larger models only when needed. Our work, however, discusses a simple independently trained two-model ensemble and switching rule, analyzes the conditions under which its performance varies, and finds concrete cases where a large model can be improved on, not just approximated.

\section{Analysis}
\subsection{Experiment Setup}

{\bf Model Architecture}. We use the T5 transformer encoder-decoder model \cite{raffel2019exploring} for our experiments using the T5X implementation using the SMALL (60M parameters), BASE (220M parameters), LARGE (770M parameters) and 3B (3B parameters) model variants. For each dataset, we fine-tune starting from the official pre-trained checkpoints using batch size of 128. For each model and dataset pair, we generated 10 separate fine-tuned models. We trained the models on the latest generation TPUs.

{\bf Datasets}. We have 4 classification-based text datasets and 2 sequence-to-sequence text datasets.

We first describe the classification-based datasets. (1) IMDB Reviews \cite{maas-EtAl:2011:ACL-HLT2011}, binary sentiment movie analysis which contains 25000 training examples and 25000 testing examples and the task is to classify whether the movie review was positive. We fine-tune for 20,000 steps. (2) Wikipedia Toxicity Subtypes  \cite{10.1145/3038912.3052591}, which comes from Wikipedia talk page comments annotated for toxicity containing 159,571 training examples and 63,978 testing examples. We use it as a binary classification task, where comments with toxicity score at least 0.01 are labeled as toxic. We fine-tune for 50,000 steps. (3) Amazon US Customer Reviews dataset, restricted to 1,758,997 reviews of video games using a random split of 80\% train and 20\% test. We use as input the review body and the task is to classify the rating (5 classes). We fine-tune for 100,000 steps. (4) Yahoo! Answers Topic Classification Dataset \cite{zhang2015character}, where the input to the model is the title, question and answer, and the task is to classify which of the 10 topics it belongs to. The dataset contains 1,400,000 training examples and 60,000 test examples. We fine-tune for 100,000 steps. 

Now we describe the sequence-to-sequence datasets. (1) Stanford Question Answering Dataset (SQuAD)  \cite{rajpurkar2016squad} benchmark dataset consisting of reading comprehension crowdsourced questions based on Wikipedia articles, where the answer to each question is a subset of that text or the question can be unanswerable. The dataset consists of 87,599 training examples and 10,570 testing examples. (2) WMT Translate \cite{wmt19translate}. We use the English-German translation task from WMT T2T \cite{bojar-EtAl:2014:W14-33} containing 4,592,289 training examples and 3,000 testing examples. We fine-tune for 250,000 steps. For this dataset alone, we do not use accuracy as a metric; instead we use the SacreBLEU implementation of BLEU, a corpus-level metric that is not appropriate to apply at the level of individual examples \cite{papineni2002bleu, post2018sacrebleu}. For that reason, we include it in only a subset of our analyses.

{\bf Model Inference}. For the classification-based text datasets, the T5 model has a decoder, the output is still a sequence which is the class label (i.e. toxic or not toxic and positive or negative). Thus, in order to obtain the class probability scores for each example, we take the likelihood scores of each class label implied by the decoder and apply a softmax to obtain the probability distribution over the classes. We fine-tuned for 250,000 steps. 

{\bf Notion of uncertainty}. For the classification based datasets, we use the probability scores for each class described above. For the sequence-to-sequence datasets, we use the average margin scores for each token in the sequence output, where to get the output of the $t$-th token, we input the correct first $t-1$ tokens to the decoder in order to obtain the $t$-th token's softmax probability distribution over the entire token value vocabulary (this procedure is also known as teacher forcing \cite{lamb2016professor}).

\subsection{Uncertainty and Accuracy}

\begin{figure}
\small
\centering
\begin{tabular}{cc}
\includegraphics[width=2in]{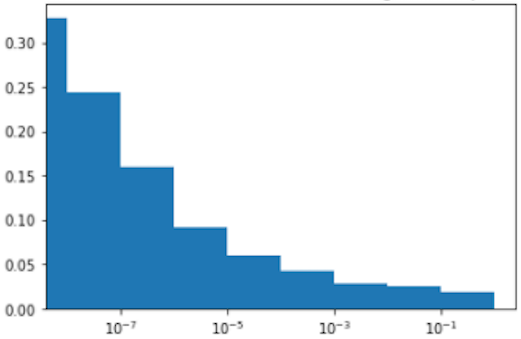}  &
\includegraphics[width=2in]{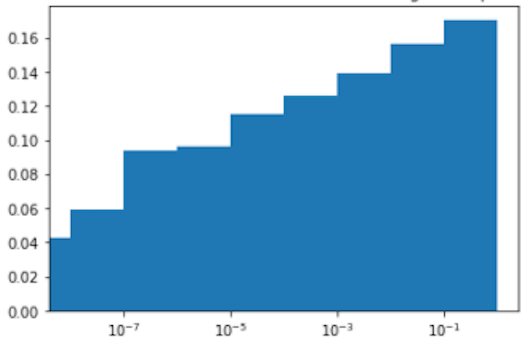} \\
a) SMALL Correct & b) SMALL Wrong \\
\includegraphics[width=2in]{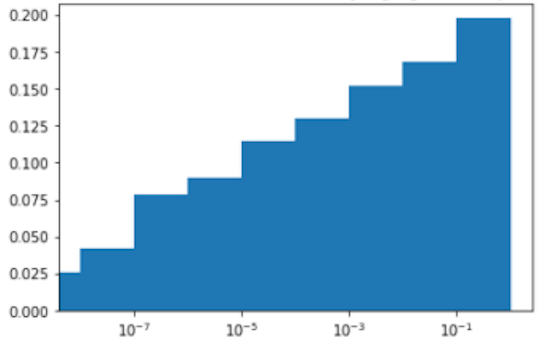} & 
\includegraphics[width=2in]{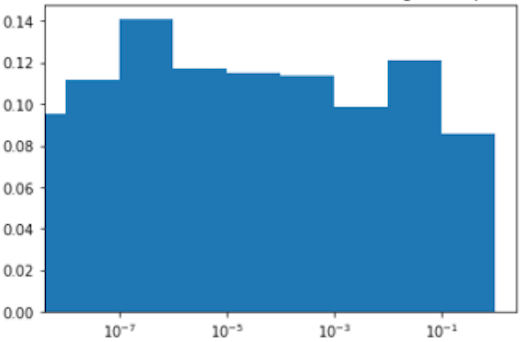} \\
c) Only 3B Correct & d) Both Wrong   \\
\end{tabular}
\caption{Histogram of distribution of predicted uncertainties among examples that a SMALL model predicted right and predicted wrong on the IMDB dataset. X-axis is the implied probability of second most likely prediction, in log-scale.} \label{fig:buckets}
\end{figure}

{\bf Big models find it easiest to improve on the smaller model's most uncertain examples}. We start with the well-known finding that a model's uncertainty scores are informative about its own performance. Namely, models are more accurate on those examples they are certain about, while they are more inaccurate on examples they are uncertain about (even if the specific softmax scores may not be well-calibrated \cite{guo2017calibration}). We see an illustration of this in Figure~\ref{fig:buckets}. Among examples that the model gets correct, it tends to have high certainty, with monotonically fewer examples in each higher-uncertainty bucket. The opposite is true in cases where the model is wrong.

It may seem like this chain of reasoning is leading to a banal conclusion - that larger models can improve more on examples a smaller model is uncertain about because that is where the smaller model tends to make mistakes.

It turns out, though, that this is not the full story. Figure~\ref{fig:buckets} also splits out the distribution of uncertainties for those examples the original model gets wrong into those that a larger model also gets wrong and those that the larger model gets right. The two histograms look nothing alike. The examples on which the 3B model improves are concentrated on the most uncertain examples, while the examples on which the 3B model is also wrong are found across the spectrum of uncertainties. It is certainly not the case that the larger model is evenly improving among those examples where there is room to improve; it is finding the most uncertain examples to be an easier target.

\begin{figure}
\small
\centering
\begin{tabular}{cc}
\includegraphics[width=2in]{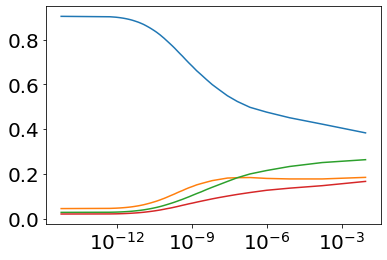}  &
\includegraphics[width=2in]{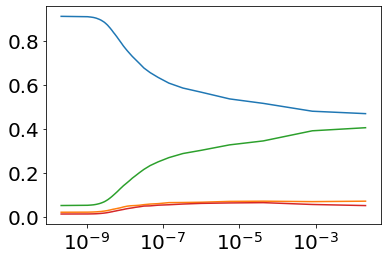} \\
a) Wikipedia & b) IMDb \\
\includegraphics[width=2in]{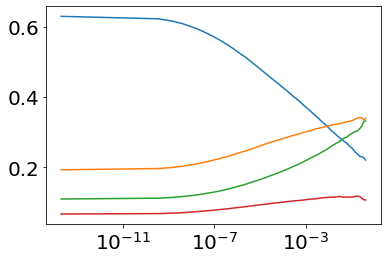} & 
\includegraphics[width=2in]{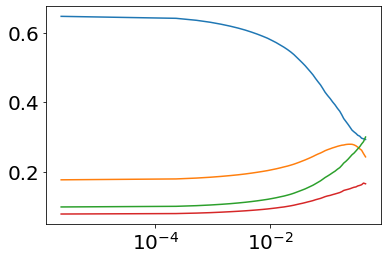} \\
c) Yahoo! & d) Amazon   \\
\includegraphics[width=2in]{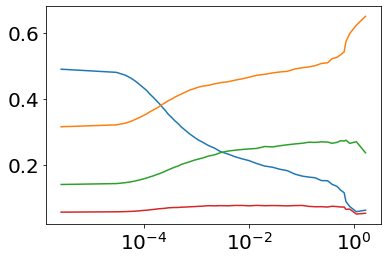} &  \\
e) SQuAD &   \\
\end{tabular}
\caption{Fraction of examples that lie in each of the 4 accuracy buckets (blue: both correct, orange: both wrong, green: only 3B model correct, red: only SMALL model correct), as we limit ourselves to examples at least as uncertain as each point on the x-axis. For example, if blue curve passes through a point (1e-8, 0.8), that indicates 80\% of examples with uncertainty smaller than 1e-8 are classified correctly by both models.}
\label{fig:bucket_line}
\end{figure}

Another way to visualize these findings is by dividing examples into four buckets, based on whether the smaller and larger model correctly classified them, and then looking at the relative size of these buckets as we limit ourselves to more uncertain examples. As we see in Figure~\ref{fig:bucket_line}, the green line - the fraction of examples only the 3B model gets right - steadily grows as we perform this subsetting. By the time we are limiting ourselves to the most uncertain examples, a strong plurality or even majority of examples are those in that specific bucket.

Note that for many datasets, the other categories in which at least one model is wrong - those in which only the SMALL model is right or both models are wrong - are also rising as a fraction of all examples as we consider the most uncertain examples. They are - in every case - rising more slowly, however.



{\bf There are uncertainty ranges where larger models improve, stagnate, and even do worse than smaller models.} We now proceed to a more direct analysis of how, at an example level, a model's uncertainty interacts with its accuracy and that of larger models. To do so, we divide examples into 100 uncertainty buckets based on a model's uncertainty scores and then calculate accuracies within each bucket of that model, both for that model as well as larger models of various sizes.

\begin{figure}
\small
\centering
\begin{tabular}{cc}
\includegraphics[width=2in]{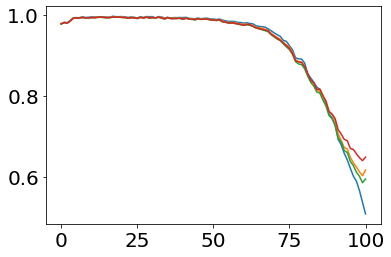}  &
\includegraphics[width=2in]{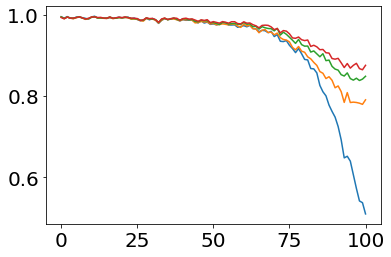} \\
a) Wikipedia & b) IMDb \\
\includegraphics[width=2in]{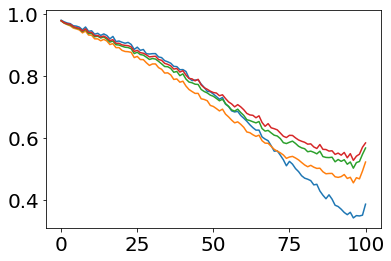} & 
\includegraphics[width=2in]{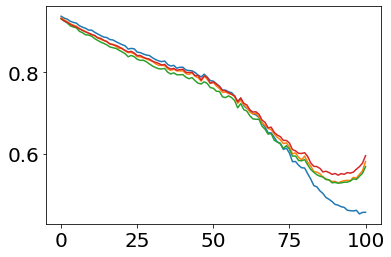} \\
c) Yahoo! & d) Amazon   \\
\includegraphics[width=2in]{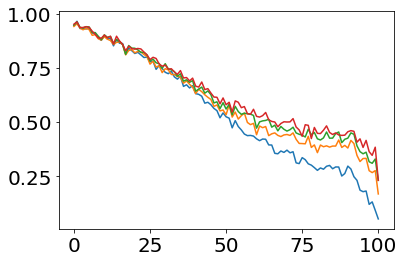} & 
\includegraphics[width=2in]{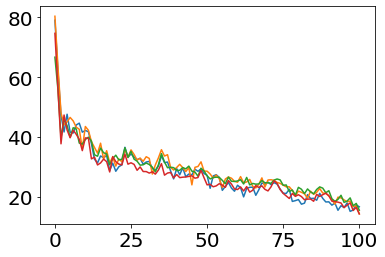} \\
e) SQuAD & f) WMT   \\
\end{tabular}
\caption{Model accuracies (BLEU score for the WMT data) by percentile of example uncertainty: blue: SMALL model, orange: BASE model, green: LARGE model, red: 3B model.}
\label{fig:accuracy_line}
\end{figure}

The results, shown in Figure~\ref{fig:accuracy_line}, are stark. In each case, there is effectively no improvement over the SMALL model - for any model size - on a significant fraction of examples; the SMALL model is the best you can do. Only as we get to the most uncertain examples do we see a gap open up between the models.

Furthermore, within each dataset, this divergence point is fairly similar for models of all sizes. The largest models do not make significantly more inroads into moderately uncertain examples relative to medium-sized models; they mostly just do ever better on the most uncertain examples.

These findings clash with some natural intuitions one might have about how model size would interact with performance. For one, smaller models retain parity with larger models well into the regime where they start to make lots of mistakes. For example, on the Yahoo dataset (bottom-right of Figure~\ref{fig:accuracy_line}), there does not start to be a difference between the models until the SMALL model dips below around 75\% accuracy. 

Second, we do not see a waterfall-like pattern where the SMALL model's accuracy is ``good enough'' for a short amount of time before falling off, after which the next largest model remains ``good enough'' for a while, and so on. This finding can have implications for model design, suggesting that ``intermediate''-sized models may not have a particularly useful niche in some contexts - examples are better handled either by a smaller model or a bigger model.

There are, of course, differences between the graphs as well across datasets. The divergence point comes at different uncertainty levels, and the improvement of larger models is therefore more or less concentrated among the most uncertain examples. More tantalizingly, the smaller models appear to actually do better than larger models among low-uncertainty examples in some of the datasets! We now turn to systematically analyzing these differences.

\subsection{Switcher Models}
\label{sec:switchers}

{\bf Switcher model curves visualize how much performance we can gain by toggling between a small and large model based on uncertainty.}
A natural way to visualize and further evaluate the significance of our findings regarding model uncertainty and larger model improvement is with the use of switcher models. Switcher models, for the purpose of this paper, consist of a two-model ensemble. The smaller model always evaluates an example. If it is uncertain about an example, it passes the example to the larger model. Switcher models of different kinds have recently been explored in the context of distillation and multi-model cascades \cite{rawat2021summon, wang2018idk}.

\begin{figure}
\small
\centering
\begin{tabular}{cc}
\includegraphics[width=2in]{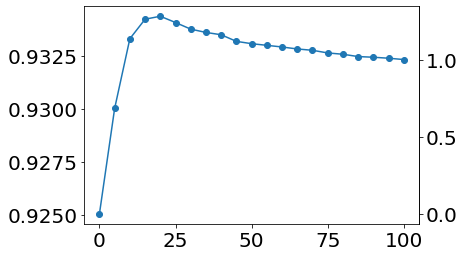}  &
\includegraphics[width=2in]{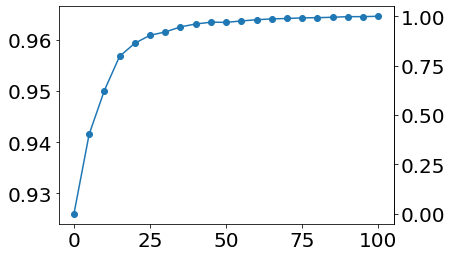} \\
a) Wikipedia & b) IMDb \\
\includegraphics[width=2in]{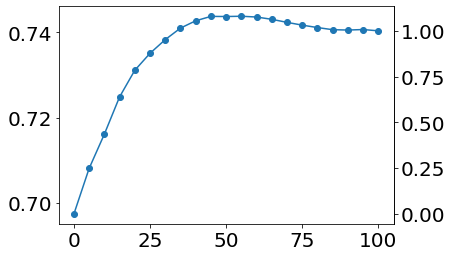} & 
\includegraphics[width=2in]{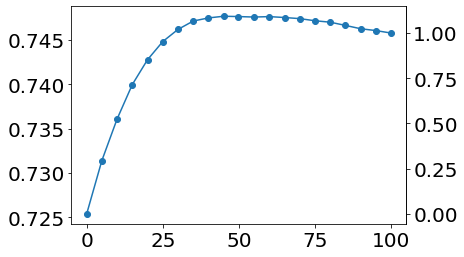} \\
c) Yahoo! & d) Amazon   \\
\includegraphics[width=2in]{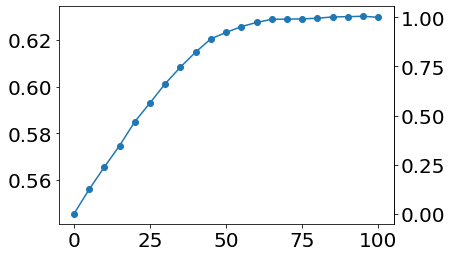} & 
\includegraphics[width=2in]{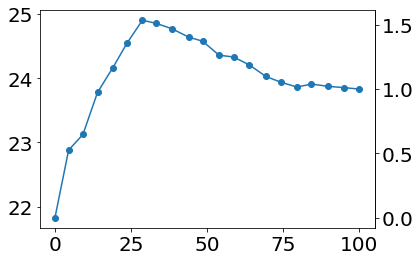} \\
e) SQuAD & f) WMT   \\
\end{tabular}
\caption{Model accuracies of switcher model that send the given fraction of the most uncertain examples based on the SMALL model to the 3B model and take the rest with the SMALL model themselves. X-axis is the proportion of examples evaluated by the 3B model; y-axis shows the accuracy (BLEU score for the WMT data), with the left axis showing the absolute accuracy and right axis showing the scaled accuracy (with the accuracy of evaluating all examples by the 3B model equaling to 1).}
\label{fig:switchers}
\end{figure}

In particular, we focus on the curve of all possible switcher models for each SMALL-LARGE pair. Consider Figure~\ref{fig:switchers}, which displays example switcher models between the SMALL and 3B T5 variant on each dataset under consideration. Each point on a switcher model plot represents the performance of a hybrid model where the 3B model is deputized on the given fraction of examples the SMALL model is most uncertain about, while the SMALL model handles the rest. The leftmost point is therefore the case where the SMALL model handles every example, while the rightmost point is where the 3B model handles every example, and the resulting switcher model accuracies at those points are equivalent to the performance of the SMALL and 3B model, respectively.

In the top-right example, we observe that while the SMALL model achieves only 93\% accuracy on its own, we can achieve a hybrid model with 95\% accuracy by sending only 10\% of examples to the 3B model. It can achieve north of 96\% accuracy, essentially equivalent to the accuracy of the big model, by sending off only half of examples.

With that in mind, we are in a position to observe just how stark the results from these switcher models are.
Despite how simple the idea is, we see a rather attractive trade-off between performance and compute in dataset after dataset. In cases where evaluating the large model in every case is expensive or otherwise underesirable, a simple switcher model can capture all or nearly all of the performance while saving substantial compute resources.

Even more interestingly, we see multiple cases where the hybrid switcher model does \emph{better} than the big model! Take the switcher model on the WMT dataset (bottom-right of Figure~\ref{fig:switchers}). By sending 35\% of the examples the SMALL model is most uncertain about to the 3B model, the hybrid model achieves a BLEU score of 25, which is well above the score of 24 achieved by the big model. For the Wikipedia dataset (top-left), we can already beat the performance of the 3B model by sending it only 15\% of examples to evaluate (and do even better by sending it 25\%).

{\bf Switcher model humps occur when a switcher model outperforms the larger model.} We refer to this phenomenon as a \emph{hump}. The existence of a hump on the switcher model is a reflection of two factors: first, the strong improvement by a larger model over the SMALL model on examples the SMALL model is uncertain about (leading to the steep rise at the beginning); and second, the decreased performance by the larger model on examples a SMALL model is certain about (leading to the decline at the end).

Even if we don't always see a hump, the first of those factors holds in all of our datasets, which results in the switcher model curve being \emph{concave}, or bowed out. The more concave a switcher model is, the more two things are true. First, we can achieve greater compute savings to achieve a desired level of performance (e.g. matching the big model). Second, we can achieve greater performance given a desired level of compute.

Concavity and humps are two things we can look out for when doing more systematic analyses of switcher models. For example, to better understand why and when switcher models are more concave or have larger humps, we look into the switcher models across model retrainings for the same dataset and architecture for SMALL and 3B models.

\begin{figure}
\centering
\includegraphics[width=3in]{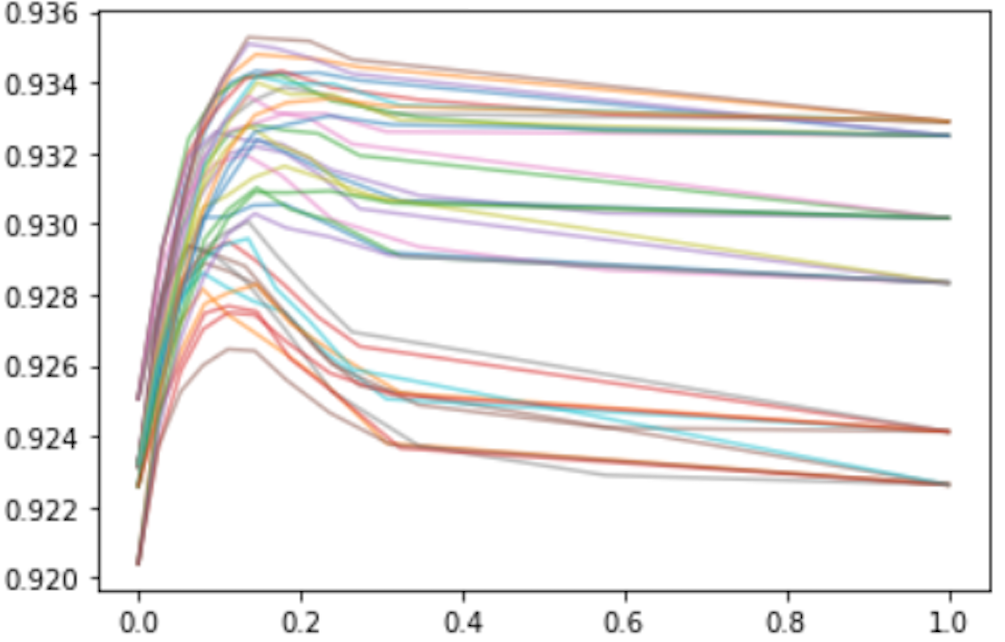}
\caption{Several switcher model curves, each corresponding to one pair of 3B and SMALL models for the Wikipedia toxicity dataset. X-axis denotes the proportion of examples evaluated by the 3B model, and y-axis is the switcher model accuracy.}
\label{fig:retrained_switcher}
\end{figure}

{\bf Switcher model humps are larger and more frequent when models are closer in accuracy and disagree on more examples.}
Figure~\ref{fig:retrained_switcher} plots several switcher models overlaid on each other for different retrains of both the 3B and SMALL model of the Wikipedia toxicity dataset. We use this dataset for our illustration as it both achieves humps and has relatively large differences in model accuracies over retraining compared to the average performance gap between the two model sizes. 

It is immediately clear that the gap between the smaller and the larger model's performance correlates strongly with the concavity of the switcher model. The closer the two models are, the more concave a switcher model will be - and the more likely it achieves a large hump. We see that for the worst-performing SMALL model and best-performing 3B model, the switcher model has a very small hump; meanwhile, the hump in the opposite case is very large.

These findings have important implications for practitioners. When a larger model has a relatively small improvement over a smaller model, there may be a substantial set of examples where it is actually performing worse. And so rather than blindly using the larger model on all examples, there may be gains to switching between the models, or ensembling more generally.

The existence of substantial humps when the larger model performs no better than the small model is also interesting. It provides some intuition for the benefits of ensembling in general; even when models perform similarly on average, they will do better or worse on different subsets and so combining their predictions in some way can improve overall performance. Here, we see that the smaller model's uncertainty is a guide to where each is doing relatively better, freeing us from having to evaluate both models in every case.

\begin{figure}
\centering
\includegraphics[width=3in]{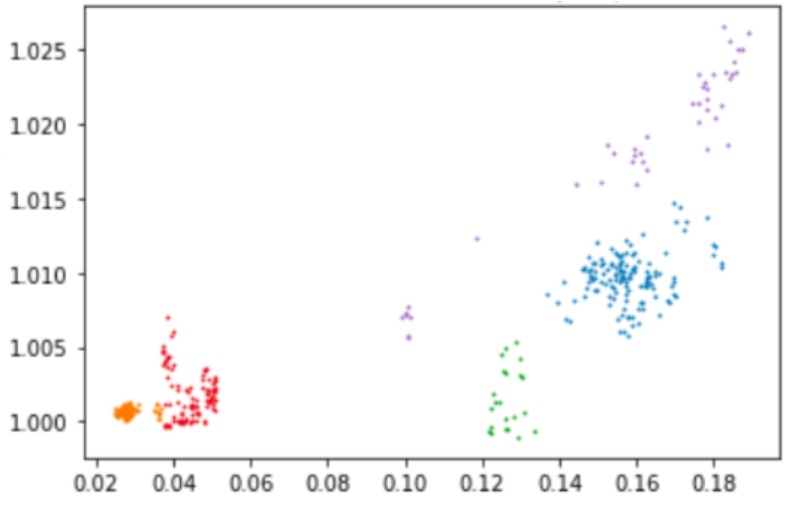}
\caption{Each point in this plot is a particular pair of a smaller and bigger model whose accuracies differ by between 0.5\% and 1\%. X-axis is the disagreement fraction, and y-axis is the switcher model hump size. Different colors of dots denotes different datasets: blue-Amazon, orange-IMDb, green-SQuAD, red-Wikipedia, purple-Yahoo!. In general, the pairs with higher churn exhibit larger switcher model humps.}
\label{fig:churn_hump}
\end{figure}

We now look across datasets. While there are not enough datasets to draw systematic conclusions about why their switcher model concavities differ (apart from different accuracy gaps), we offer suggestive evidence in Figure~\ref{fig:churn_hump} that model disagreement rates, or the fraction of examples on which a larger and smaller model disagree, is part of the story. We limit ourselves to only considering pairs of models with very similar accuracy gains to one another, between 0.5\% and 1\%, in order to limit the effect of that factor. Within the remaining group of models, those that exhibit higher disagreement relative to one another are far more likely to exhibit humps.

Intuitively, we argue that this makes sense. Greater disagreement, when holding fixed the difference in accuracies, means that each model wins relative to the other more often. If these wins are concentrated in the more and less uncertain examples, respectively, the result will be more concavity and a larger hump in the switcher model.



\subsection{Different Notions of Uncertainty}
\label{sec:uncertainties}

{\bf A small model's uncertainty is a better predictor of large model performance than a committee's uncertainty, but its rate of disagreement with a committee of small models is even better.} Besides differences between datasets and models, one other factor has the ability to strongly influence a switcher model's concavity - the type of uncertainty score being used. Are some uncertainty scores better than others at distinguishing which examples a larger model is likely to get right or wrong?

In particular, we consider the question of whether uncertainty scores that are specific to an individual small model, such as the margin, are better predictors of a larger model's improvement than committee-based uncertainty scores that rely on evaluating multiple models. Committee-based scores clearly bring more information to the table on the whole; at the same time, an individual model's uncertainty has better information about that model's own strengths and weaknesses. It has also has the advantage of being more practical and useful; if we wanted to use a switcher model, we only need a small model and a big model, not an ensemble of other small models to help decide which examples to forward to the big model.

For committee scores, we consider both averaging uncertainty across small models and also accounting for how much retraining churn there is for each example, i.e. how often models have different predictions. Retraining churn has shown some interesting correlations with switcher model performance in Section~\ref{sec:switchers}, and is also intuitively related to the idea of uncertainty. Table~\ref{table:uncertainty-churn} in the Appendix shows that it is quite correlated with a model's uncertainty score, with much more churn on examples a model is uncertain about, but not equivalent. For example, there is a fair amount of churn on examples in the most certain bucket.

\begin{table*}
\small
\centering
\begin{tabular}{llcccccc}
\toprule
Dataset & Larger Model & Margin  & Cmt. Margin  & Entropy & Cmt. Entropy  & Churn & Cmt. Churn       \\
\midrule
Wikipedia  & BASE & 0.2\% (0.01\%) & 0.1\% (0.01\%) & 0.2\% (0.01\%) & 0.1\% (0.01\%) & 0.3\% (0.01\%) & 0.1\% (0.01\%) \\
Wikipedia &  LARGE & 0.2\% (0.01\%) & 0.1\% (0.01\%) & 0.2\% (0.01\%) & 0.1\% (0.01\%) & 0.3\% (0.02\%) & 0.1\% (0.01\%) \\
Wikipedia & 3B & 0.3\% (0.02\%) & 0.2\% (0.01\%) & 0.3\% (0.02\%) & 0.2\% (0.01\%) & 0.3\% (0.01\%) & 0.1\% (0.01\%) \\
\midrule
IMDb & BASE & 0.0\% (0.00\%) & 0.1\% (0.00\%) & 0.0\% (0.00\%) & 0.1\% (0.00\%) & 0.0\% (0.00\%) & 0.0\% (0.00\%) \\
IMDb & LARGE & 0.0\% (0.00\%) & 0.0\% (0.00\%) & 0.0\% (0.00\%) & 0.0\% (0.00\%) & -0.0\% (0.00\%) & -0.0\% (0.00\%) \\
IMDb & 3B & 0.0\% (0.00\%) & 0.0\% (0.00\%) & 0.0\% (0.00\%) & 0.0\% (0.00\%) & -0.0\% (0.00\%) & -0.0\% (0.00\%)\\
\midrule
Yahoo! & BASE & 2.1\% (0.06\%) & 0.6\% (0.04\%) & 2.1\% (0.06\%) & 0.7\% (0.04\%) & 4.2\% (0.10\%) & 0.8\% (0.05\%) \\
Yahoo! & LARGE & 1.1\% (0.10\%) & 0.4\% (0.04\%) & 1.1\% (0.10\%) & 0.4\% (0.05\%) & 2.3\% (0.16\%) & 0.4\% (0.06\%) \\
Yahoo! & 3B & 0.5\% (0.03\%) & 0.1\% (0.00\%) & 0.5\% (0.03\%) & 0.1\% (0.00\%) & 1.4\% (0.05\%) & 0.0\% (0.01\%) \\
\midrule
Amazon & BASE & 1.0\% (0.03\%) & 0.4\% (0.02\%) & 1.1\% (0.03\%) & 0.3\% (0.02\%) & 2.4\% (0.03\%) & 0.7\% (0.02\%) \\
Amazon & LARGE & 0.9\% (0.04\%) & 0.5\% (0.02\%) & 0.9\% (0.04\%) & 0.4\% (0.02\%) & 2.3\% (0.05\%) & 0.9\% (0.03\%) \\
Amazon & 3B & 0.5\% (0.03\%) & 0.3\% (0.02\%) & 0.5\% (0.03\%) & 0.3\% (0.02\%) & 1.8\% (0.04\%) & 0.7\% (0.03\%) \\
\midrule
SQuAD & BASE & 0.2\% (0.01\%) & 0.1\% (0.01\%) & 0.2\% (0.01\%) & 0.1\% (0.01\%) & 0.6\% (0.03\%) & 0.0\% (0.01\%) \\
SQuAD & LARGE & 0.1\% (0.01\%) & 0.1\% (0.01\%) & 0.0\% (0.01\%) & 0.0\% (0.00\%) & 0.0\% (0.01\%) & -0.0\% (0.01\%) \\
SQuAD & 3B & 0.1\% (0.01\%) & 0.1\% (0.01\%) & 0.1\% (0.01\%) & 0.1\% (0.01\%) & 0.0\% (0.01\%) & -0.0\% (0.01\%) \\
\bottomrule
\end{tabular}
\caption{Average percentage improvement in accuracy at the peak of the hump compared to the maximum accuracy of the SMALL and larger models. We consider three variants of uncertainty scores: margin, entropy, and churn (i.e., the probability that the given SMALL model makes a different prediction than another finetuning of the SMALL model). We also consider the committee-based variants of the uncertainty scores, which use the average of uncertainties of nine other finetunings of the SMALL model.}
\label{table:hump-pct}
\end{table*}

The results, in Tables~\ref{table:hump-pct} and \ref{table:avg-concavity-pct} (in Appendix), are quite interesting. A model's own uncertainty score is a more useful input to a switcher model than the average uncertainty score across several retrained models. However, the most useful score in this context is actually the churn! That is, the best way to predict when a model can be improved by a larger model and when it will likely outperform that same model is by computing how often it disagrees with retrained peers.

\section{Discussion}

This study raises a lot of interesting questions that are worth further exploration. Particularly, why do we observe this relationship between regions of uncertainty in small models and improvement in performance using larger models? Also, why can larger models have worse performance on samples that are predicted with high certainty by small models? And what about our findings about individual model uncertainty scores, average uncertainty scores, and churn?

Let us start with the first question. Why is that, given a set of examples that a smaller model gets wrong, bigger models tend to mostly improve only on the subset where the smaller model is most uncertain? Note that this exists somewhat in tension with the conventional idea that uncertain examples are ``hard'' and certain examples are ``easy''. Indeed, if we think of example hardness as expressing how likely it is that even a larger and more complex model would struggle, then this suggests that a fraction of a model's most uncertain examples are actually the next easiest!

At the same time, in aggregate, larger models still perform worse on those examples a small model is uncertain about than on those it is certain about - even after their improvement on that slice - as we see in Figure~\ref{fig:accuracy_line} where all lines slope downward. How can we square these two observations?

The answer likely lies in the many reasons why a small model can be uncertain. One reason it may be uncertain because an example is genuinely challenging. For example, an example may be out of distribution - i.e. very different from examples in the training set. In that case, a larger model will also struggle. Another reason, though, could be because of model undercapacity; it simply lacks the flexibility to navigate complex parts of the feature space and defaults to having high-uncertainty predictions in that region. These would be prime examples of where a larger model can improve, as it can make more fine changes in the decision boundary to classify more examples correctly. Because a model's uncertainty does not distinguish between these (and other) reasons for uncertainty, we end up with the surprising result that those examples are both ``easy'' (a larger model can get many of them right) and ``hard'' (all models are worse on average).


The flip question is why it is that smaller models can sometimes outperform larger models, on average, on examples they are highly certain about. 

At a high level, the most common reason why smaller models outperform larger models is because of overfitting. How could that play out in this context, when we only observe large model underperformance on some examples the small model is quite certain about?

Consider, for example, the case of label noise - a point from one class surrounded by points from another. A small model may choose to ignore the anomalous point, and therefore be highly certain about its predictions on all points in the region. A larger model, meanwhile, could contort itself in an unwieldy manner and thereby generalize poorly to a test set. More generally, given the extreme high-dimensionality of the feature space, it's plausible that there are regions where simple decision boundaries yield high-certainty high-accuracy classifications.

We now discuss the findings about the various scores to rank the examples for the switcher model. One of the insights was that the small model's own uncertainty was a better score than the uncertainty of the committee of small models, which is known to be a stronger uncertainty estimator. One reason why this is the case is that the models' own uncertainty is more relevant to that model than a committee. A committee's most certain examples could also be those that are so easy that a larger model is unlikely to underperform, limiting the ability for a switcher model to outperform the larger model. This suggests that there is added predictive power in the committee and using the disagreement is an effective way to both leverage the power of the committee while still basing the score on the model's own predictions. It also suggests that large models are particularly vulnerable to worsening in performance in cases where a large set of small models all agree.

\section{Conclusions}

We identified several regularities in the relationship between a model's example-level uncertainty and the performance of it and larger models. Across several datasets, we found that larger models improve the most on the most uncertain subset of examples, and don't improve at all on a large fraction of examples. We also found that in cases where the large models do not improve by much or differ from the smaller model on many examples, the smaller model can outperform the larger model on an identifiable subset of examples. All of these correlations become even stronger, meanwhile, when we consider how often a smaller model disagrees with other small model retrainings. 


We hope that this paper spurs further work in understanding when increasingly large models are actually needed, and when we can save the compute or even increase performance by using a smaller model. 

\bibliography{example_paper}
\bibliographystyle{abbrvnat}


\newpage
\appendix
\onecolumn

\section{Additional Experiments}

\begin{table*}[!ht]
\small
\centering
\begin{tabular}{llccccc}
\toprule
Dataset & Uncertainty Model & \% Churn Q1  & \% Churn Q2  & \% Churn Q3  & \% Churn Q4  & \% Churn Q5        \\
\midrule
Wikipedia & SMALL & 0.41\% (0.00\%)  &  1.42\% (0.00\%)  &  6.22\% (0.01\%)  &  16.40\% (0.02\%)  &  20.63\% (0.08\%) \\
Wikipedia & BASE & 0.79\% (0.00\%)  &  0.91\% (0.00\%)  &  4.45\% (0.01\%)  &  18.34\% (0.03\%)  &  23.19\% (0.08\%) \\
Wikipedia & LARGE & 0.50\% (0.00\%)  &  0.86\% (0.00\%)  &  4.61\% (0.01\%)  &  19.42\% (0.03\%)  &  26.67\% (0.09\%) \\
Wikipedia & 3B & 0.16\% (0.00\%)  &  0.79\% (0.00\%)  &  4.44\% (0.01\%)  &  20.94\% (0.03\%)  &  34.81\% (0.09\%) \\
\midrule
IMDb & SMALL & 0.33\% (0.00\%)  &  1.26\% (0.00\%)  &  4.36\% (0.02\%)  &  7.78\% (0.03\%)  &  8.08\% (0.09\%) \\
IMDb & BASE & 0.35\% (0.00\%)  &  0.70\% (0.00\%)  &  2.73\% (0.01\%)  &  11.21\% (0.03\%)  &  14.58\% (0.11\%) \\
IMDb & LARGE & 0.14\% (0.00\%)  &  0.47\% (0.00\%)  &  2.20\% (0.01\%)  &  13.22\% (0.04\%)  &  20.77\% (0.13\%) \\
IMDb & 3B & 0.20\% (0.00\%)  &  0.40\% (0.00\%)  &  1.31\% (0.01\%)  &  13.23\% (0.04\%)  &  32.44\% (0.15\%) \\
\midrule
Yahoo! & SMALL & 7.00\% (0.01\%)  &  23.54\% (0.01\%)  &  33.10\% (0.02\%)  &  36.00\% (0.03\%)  &  33.37\% (0.10\%) \\
Yahoo! & BASE & 7.20\% (0.01\%)  &  24.06\% (0.01\%)  &  32.29\% (0.02\%)  &  34.88\% (0.03\%)  &  32.69\% (0.10\%) \\
Yahoo! & LARGE & 5.94\% (0.01\%)  &  22.14\% (0.01\%)  &  35.63\% (0.03\%)  &  40.86\% (0.03\%)  &  39.73\% (0.10\%) \\
Yahoo! & 3B & 3.99\% (0.01\%)  &  21.42\% (0.01\%)  &  39.44\% (0.03\%)  &  46.82\% (0.03\%)  &  44.61\% (0.10\%) \\
\midrule
Amazon & SMALL & 8.31\% (0.00\%)  &  18.03\% (0.00\%)  &  27.69\% (0.01\%)  &  31.08\% (0.01\%)  &  28.77\% (0.04\%) \\
Amazon & BASE & 8.24\% (0.00\%)  &  18.77\% (0.00\%)  &  26.24\% (0.01\%)  &  31.68\% (0.01\%)  &  30.06\% (0.04\%)\\
Amazon & LARGE & 7.09\% (0.00\%)  &  21.31\% (0.00\%)  &  24.92\% (0.01\%)  &  32.93\% (0.01\%)  &  32.65\% (0.04\%)\\
Amazon & 3B & 5.59\% (0.00\%)  &  21.58\% (0.00\%)  &  26.56\% (0.01\%)  &  37.25\% (0.01\%)  &  37.30\% (0.04\%)\\
\midrule
SQuAD & SMALL & 8.97\% (0.02\%)  &  27.54\% (0.04\%)  &  35.38\% (0.07\%)  &  34.55\% (0.09\%)  &  36.70\% (0.28\%) \\
SQuAD & BASE & 7.44\% (0.02\%)  &  26.37\% (0.04\%)  &  38.29\% (0.07\%)  &  40.75\% (0.09\%)  &  42.56\% (0.28\%)\\
SQuAD & LARGE & 6.27\% (0.02\%)  &  25.60\% (0.04\%)  &  40.89\% (0.07\%)  &  44.49\% (0.09\%)  &  47.75\% (0.29\%)\\
SQuAD & 3B & 5.73\% (0.02\%)  &  24.81\% (0.04\%)  &  42.00\% (0.07\%)  &  47.27\% (0.10\%)  &  52.98\% (0.29\%)\\
\bottomrule
\end{tabular}
\caption{Percentage of chance (standard error) that two runs of finetuning of 3B models will give different classifications on test examples. Test examples are divided into five buckets based on their uncertainty scores using the uncertainty model. Q1: 50\% most certain test examples. Q2: Test examples with uncertainty score in the 50\% to 75\% quantiles. Q3: Test examples with uncertainty score in the 75\% to 90\% quantiles. Q4: Test examples with uncertainty score in the 90\% to 99\% quantiles. Q5: 1\% most uncertain test examples.}
\label{table:uncertainty-churn}
\end{table*}

\begin{table*}[!ht]
\small
\centering
\begin{tabular}{llcccccc}
\toprule
Dataset & Larger Model & Margin  & Cmt. Margin  & Entropy & Cmt. Entropy  & Churn & Cmt. Churn     \\
\midrule
Wikipedia & BASE & 0.2\% (0.01\%) & 0.1\% (0.02\%) & 0.2\% (0.01\%) & 0.1\% (0.02\%) & 0.2\% (0.01\%) & 0.2\% (0.01\%) \\
Wikipedia & LARGE & 0.1\% (0.02\%) & -0.0\% (0.02\%) & 0.1\% (0.02\%) & -0.0\% (0.02\%) & 0.2\% (0.02\%) & 0.1\% (0.01\%) \\
Wikipedia & 3B & 0.3\% (0.01\%) & 0.2\% (0.02\%) & 0.3\% (0.01\%) & 0.2\% (0.02\%) & 0.3\% (0.01\%) & 0.2\% (0.01\%)\\
\midrule
IMDb & BASE & 1.0\% (0.01\%) & 1.0\% (0.01\%) & 1.0\% (0.01\%) & 1.0\% (0.01\%) & 0.9\% (0.01\%) & 0.9\% (0.01\%) \\
IMDb & LARGE & 1.4\% (0.01\%) & 1.5\% (0.01\%) & 1.4\% (0.01\%) & 1.5\% (0.01\%) & 1.2\% (0.01\%) & 1.1\% (0.01\%) \\
IMDb & 3B & 1.7\% (0.01\%) & 1.7\% (0.01\%) & 1.7\% (0.01\%) & 1.7\% (0.01\%) & 1.3\% (0.01\%) & 1.2\% (0.01\%) \\
\midrule
Yahoo! & BASE & 1.5\% (0.03\%) & 0.6\% (0.04\%) & 1.5\% (0.03\%) & 0.6\% (0.04\%) & 2.4\% (0.05\%) & 0.8\% (0.04\%) \\
Yahoo! & LARGE & 2.1\% (0.05\%) & 1.4\% (0.05\%) & 2.1\% (0.05\%) & 1.4\% (0.06\%) & 3.0\% (0.06\%) & 1.5\% (0.05\%) \\
Yahoo! & 3B & 2.4\% (0.03\%) & 1.7\% (0.04\%) & 2.4\% (0.03\%) & 1.8\% (0.04\%) & 3.3\% (0.05\%) & 1.8\% (0.03\%) \\
\midrule
Amazon & BASE & 0.8\% (0.01\%) & 0.4\% (0.01\%) & 0.8\% (0.01\%) & 0.3\% (0.01\%) & 1.4\% (0.01\%) & 0.6\% (0.01\%) \\
Amazon & LARGE & 1.0\% (0.01\%) & 0.6\% (0.01\%) & 1.0\% (0.01\%) & 0.6\% (0.01\%) & 1.8\% (0.01\%) & 1.0\% (0.01\%) \\
Amazon & 3B & 1.2\% (0.01\%) & 0.8\% (0.01\%) & 1.2\% (0.01\%) & 0.8\% (0.01\%) & 2.0\% (0.01\%) & 1.2\% (0.01\%) \\
\midrule
SQuAD & BASE & 2.5\% (0.02\%) & 2.1\% (0.03\%) & 2.4\% (0.02\%) & 2.1\% (0.02\%) & 3.5\% (0.02\%) & 2.3\% (0.02\%) \\
SQuAD & LARGE & 3.3\% (0.02\%) & 3.0\% (0.02\%) & 3.1\% (0.02\%) & 2.9\% (0.02\%) & 4.2\% (0.02\%) & 3.1\% (0.02\%) \\
SQuAD & 3B & 3.6\% (0.02\%) & 3.4\% (0.02\%) & 3.5\% (0.02\%) & 3.3\% (0.02\%) & 4.6\% (0.03\%) & 3.5\% (0.02\%) \\
\bottomrule
\end{tabular}
\caption{Average percentage improvement in accuracy of sending most uncertain examples to the larger model, compared to the alternative of sending a same proportion of randomly selected examples to the larger model. We consider three variants of uncertainty scores: margin, entropy, and churn (i.e., the probability that the given SMALL model makes a different prediction than another finetuning of the SMALL model). We also consider the committee-based variants of the uncertainty scores, which use the average of uncertainties of nine other finetunings of the SMALL model.}
\label{table:avg-concavity-pct}
\end{table*}


\end{document}